\documentclass[letterpaper, 10 pt, conference]{ieeeconf}  

\IEEEoverridecommandlockouts                              

\usepackage{hyperref}
\hypersetup{
    colorlinks=true,
    linkcolor=blue,
    filecolor=magenta,      
    urlcolor=cyan,
}

\title{\LARGE \bf
TUM-VIE: The TUM Stereo Visual-Inertial Event Dataset}

\author{Simon Klenk$^{*}$, Jason Chui$^{*}$, Nikolaus Demmel, Daniel Cremers
\thanks{$^{*}$ These authors contributed equally. The authors are with the Computer Vision Group, Department of Informatics, Technical University of Munich, Germany. 
\texttt{\{simon.klenk, jason.chui, nikolaus.demmel, cremers\}@tum.de}.
}%
}

\usepackage[sorting=none, maxbibnames=99]{biblatex}
\usepackage{amsmath}
\usepackage{graphicx}
\usepackage{caption}
\usepackage{subcaption}
\usepackage{balance}
\newcommand{\degree}{^{\circ}}
\renewcommand{\textdegree}{$\degree$}

\begin{document}

\maketitle
\thispagestyle{empty}
\pagestyle{empty}

\begin{abstract}
Event cameras are bio-inspired vision sensors which measure per pixel brightness changes. They offer numerous benefits over traditional, frame-based cameras, including low latency, high dynamic range, high temporal resolution and low power consumption. Thus, these sensors are suited for robotics and virtual reality applications. To foster the development of 3D perception and navigation algorithms with event cameras, we present the TUM-VIE dataset. It consists of a large variety of handheld and head-mounted sequences in indoor and outdoor environments, including rapid motion during sports and high dynamic range scenarios. The dataset contains stereo event data, stereo grayscale frames at 20Hz as well as IMU data at 200Hz. Timestamps between all sensors are synchronized in hardware. The event cameras contain a large sensor of 1280x720 pixels, which is significantly larger than the sensors used in existing stereo event datasets. We provide ground truth poses from a motion capture system at 120Hz during the beginning and end of each sequence, which can be used for trajectory evaluation. 
TUM-VIE includes challenging sequences where state-of-the art visual SLAM algorithms either fail or result in large drift. Hence, our dataset can help to push the boundary of future research on event-based visual-inertial perception algorithms.
\end{abstract}

\section{INTRODUCTION}
\begin{figure}[h!]
\vspace*{0.25cm}
     \centering
     \begin{subfigure}[b]{0.23\textwidth}
         \centering
         \includegraphics[width=\textwidth]{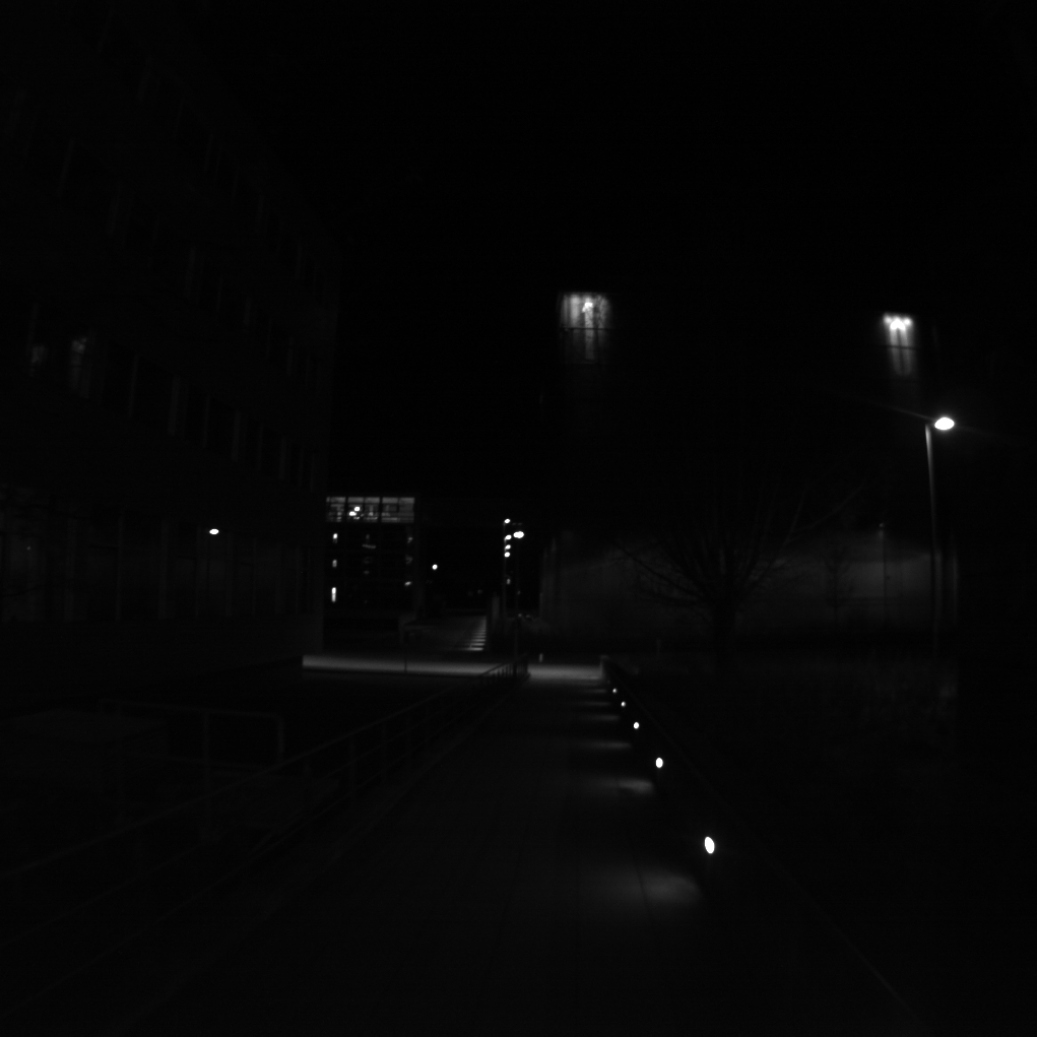}
         \caption{left visual frame}
         \label{fig:bikec0}
     \end{subfigure}
     \hfill
     \begin{subfigure}[b]{0.23\textwidth}
         \centering
         \includegraphics[width=\textwidth]{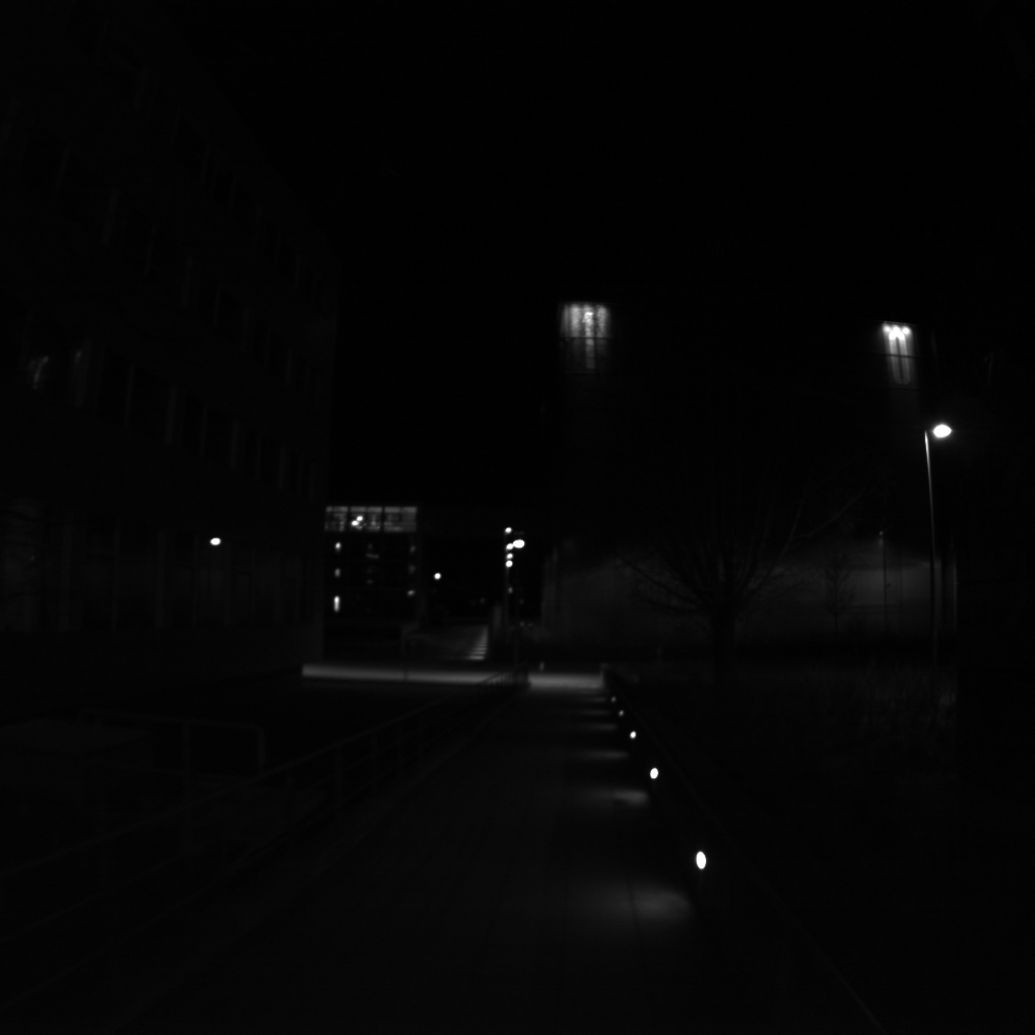}
         \caption{right visual frame}
         \label{fig:bikec1}
     \end{subfigure}
     \vskip\baselineskip
     \begin{subfigure}[b]{0.23\textwidth}
         \centering
         \includegraphics[width=\textwidth]{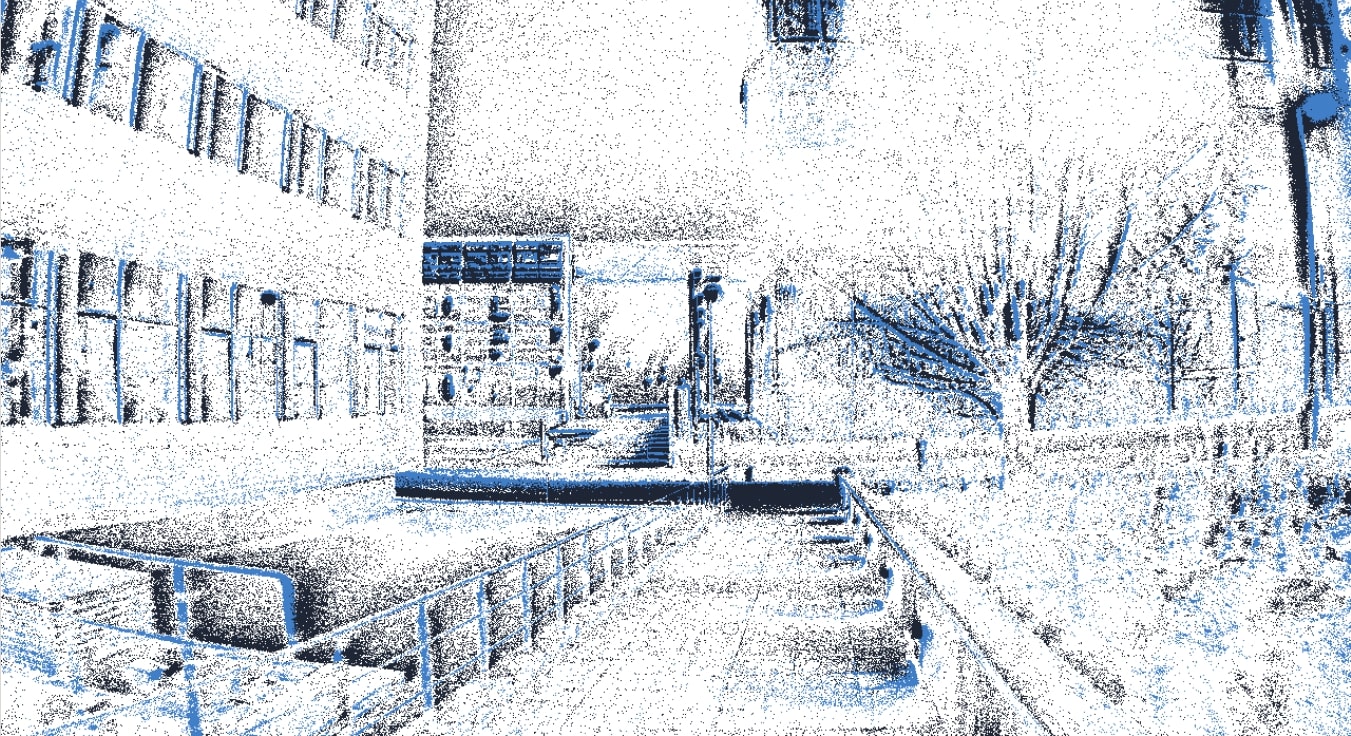}
         \caption{left event frame}
         \label{fig:bikec2}
     \end{subfigure}
     \hfill
     \begin{subfigure}[b]{0.23\textwidth}
         \centering
         \includegraphics[width=\textwidth]{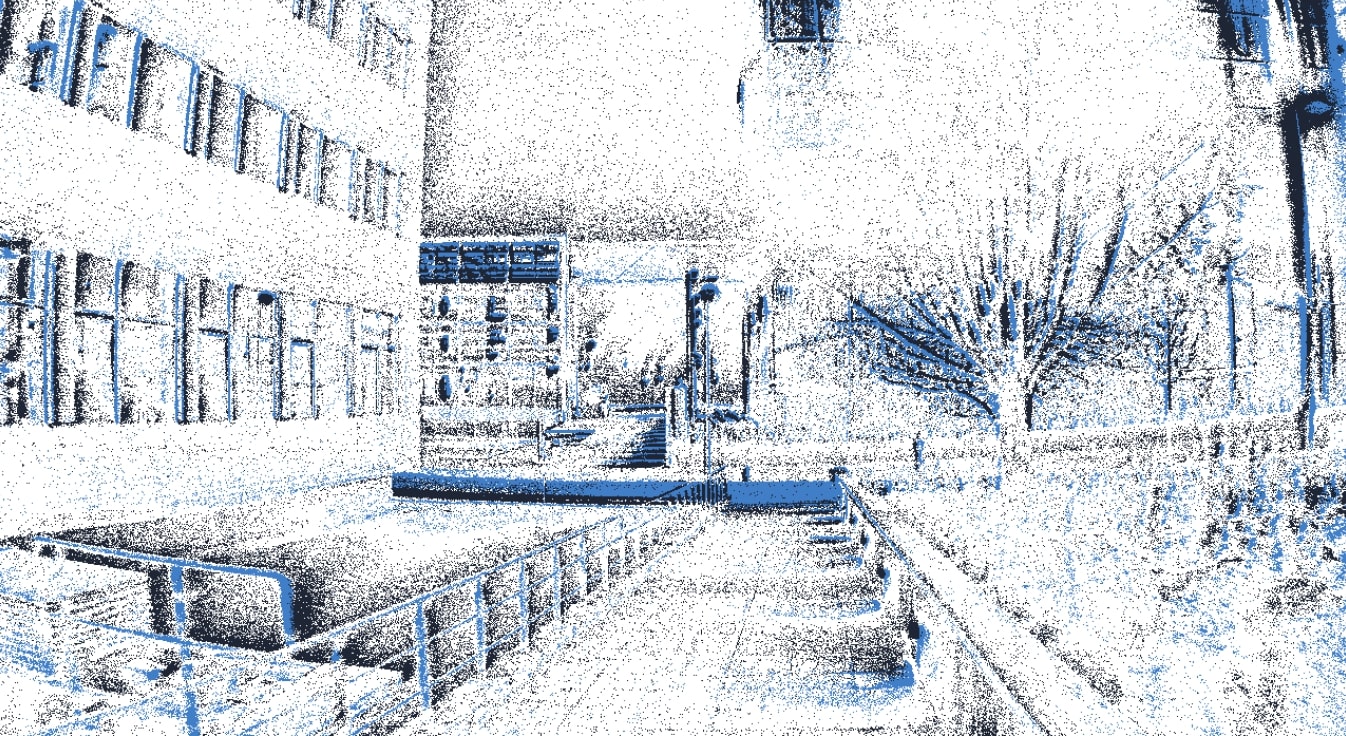}
         \caption{right event frame}
         \label{fig:bikec3}
     \end{subfigure}
        \caption{This scene is captured in low-light conditions outside. It demonstrates the high dynamic range advantage of the event camera (bottom row) compared to the visual camera (top row). It shows all four camera modalities of the sequence \textbf{bike-dark} at 83.6 seconds recording time. Note that later in this sequence there is more ambient light present and hence the visual frames can still be used in computer vision algorithms. For this visualization, the asynchronous events are accumulated into frames using an integration time of 5 milliseconds.}
        \label{fig::bikeNightScreenshots}
\end{figure}

Event cameras, also known as dynamic vision sensors (DVS), are passive imaging sensors, which report changes in the observed logarithmic brightness independently per pixel. Their main benefits are very high dynamic range (up to 140dB compared to 60dB of traditional cameras), high temporal resolution and low latency (in the order of microseconds), low power consumption and strongly reduced motion blur \cite{gallego2020Survey,delbruck2020v2e}. Hence, these novel sensors have the potential to revolutionize robotic perception. Figure \ref{fig::bikeNightScreenshots} shows the superior performance of an event camera in low light conditions.

Due to the novelty of the field, algorithms for event-based sensors are still immature compared to frame-based algorithms. Computer vision research in the last fifty years has strongly profited from publicly available datasets and benchmarks. To advance research with these novel and expensive sensors, we introduce the TUM-VIE dataset. It contains a variety of sequences captured by a calibrated and hardware-synchronized stereo pair of Prophesee GEN4 CD sensors with 1280x720 pixels resolution. To our knowledge, TUM-VIE is the first stereo dataset featuring such a high-resolution event camera, surpassing other datasets by at least a factor of ten in the number of event pixels per camera. In principle, this allows for more detailed reconstructions. More importantly, it enables the evaluation of frame-based versus event-based algorithms on data with similar, state-of-the-art resolution. To facilitate this, our setup contains two hardware-synchronized global shutter cameras of resolution 1024x1024 pixels, 12 bit color depth, known exposure times and vignette calibration. The IMU provides 3-axis accelerometer and gyroscope data at 200 Hz. Contrary to most existing datasets, we provide the calibration of IMU biases, as well as axis scaling and misalignment similar to \cite{schubert2018vidataset}. The timestamps of all sensors are synchronized in hardware.

TUM-VIE can for example be used to develop algorithms for localization and mapping (visual odometry and SLAM), feature detection and tracking, 3D reconstruction, as well as self-supervised learning and sensor fusion. The availability of comprehensive datasets combining comparable sensor modalities to TUM-VIE is quite small. Furthermore, sensor characteristics of event camera have greatly improved within the last decade \cite{gallego2020Survey}. Hence, we believe it is important to use the most recent event camera for a meaningful comparison with frame-based algorithms.

The main contributions of this paper are:
\begin{itemize}
\item We present the first 1 megapixel stereo event dataset featuring IMU data at 200Hz and stereo global shutter grayscale frames at 20Hz with known photometric calibration and known exposure times. Timestamps between all sensors are hardware-synchronized.
\item The first stereo event dataset featuring head mounted sequences (relevant for VR) and sport activities with rapid and high-speed motions (biking, running, sliding, skateboarding).
\item We propose a new method for event camera calibration using Time Surfaces.
\item We evaluate our dataset with state-of-the-art visual odometry algorithms. Furthermore, we make all calibration sequences publicly available.
\end{itemize}

\noindent The dataset can be found at:\\
\centerline{\href{https://go.vision.in.tum.de/tumvie}{https://go.vision.in.tum.de/tumvie}}

\section{Related Work}
Weikersdorfer et al. \cite{weikersdorfer2014slam} present one of the earlier event datasets, combining data from an eDVS of 128x128 pixels and an RGB-D sensor in a small number of indoor sequences. They provide ground truth poses from a motion capture (MoCap) system, but the total sequence length only amounts to 14 minutes. 

Barrancko et al. \cite{barranco2016visnav} provide a dataset which focuses on evaluation of visual navigation tasks. They use a dynamic and active pixel vision sensor (DAVIS) which combines event detection alongside regular frame-based pixels in the same sensor. However, only 5 degree-of-freedom (DOF) motions are captured and the ground truth poses are acquired from wheel odometry which is subject to drift.

The work by Mueggler et al. \cite{mueggler2016ECDS} captures full 6 DOF motions and precise ground truth by a MoCap system indoors. The sequences include artificial scenes such as geometric shapes, posters or cart boxes, but also an urban environment and an office. Hardware-synchronized grayscale frames and IMU measurements are provided in addition to the events. They also present an event camera simulator in their work. However, the monocular DAVIS 240C merely captures 240x180 pixels and the total sequence length only amounts to 20 minutes.

\begin{figure}[t]
      \centering
      \includegraphics[scale=0.49]{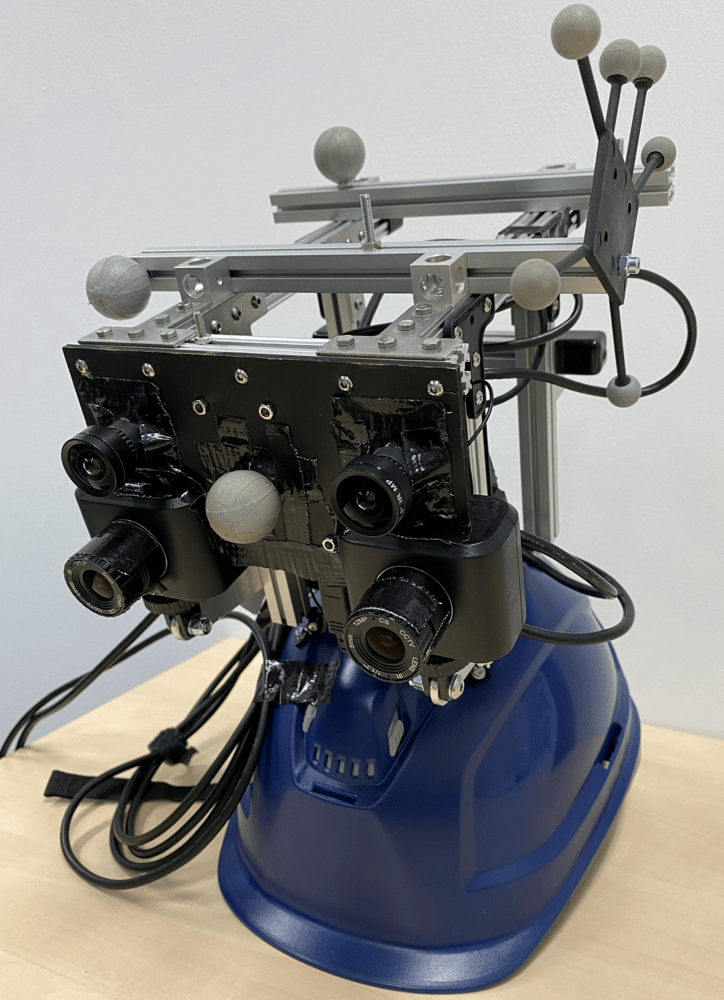}
      \caption{Sensor setup mounted onto a helmet. The distance between event camera (bottom) and visual camera (top) is only 4.49 centimeters for the left and right side of the stereo rig. The baseline between the visual cameras is 10.86 centimeters, the baseline between the event cameras is 11.84 centimeters. These numbers are obtained from our calibration algorithm and confirmed by physical measurements.}
      \label{fig::setup}
\end{figure}

There also exits a small number of related datasets targeting automated driving applications. Binas et al. present DDD17 \cite{binas17ddd17} as well as the follow-up dataset DDD20 \cite{huddd20} which feature monocular event data from a DAVIS 346B with 346x260 pixels. The camera is mounted on a car windshield while driving through various environments and conditions, for a total of 12 and 51 hours, respectively. Additionally, GPS data and vehicle diagnostic data such as steering angle and vehicle speed are provided. Similarly, the dataset by Perot et al. \cite{perot20Neurips} comprises 14 hours of recording from a car. It includes labelled bounding boxes for the classes of car, pedestrian and two-wheeler. They provide RGB images recorded at 4 megapixels as well as an event stream recorded by the 1 megapixel Prophesee GEN4-CD sensor, which is also used in this work. 

Furthermore, there exit specialised datasets such as the UZH-FPV Drone Racing Dataset \cite{Delmerico19icra}, which is targeted for localization of drones during high speed and high accelerations in 6 DOF. The sensor setup includes a miniDAVIS346 as well as a fisheye stereo camera with 640x480 pixels resolution and hardware-synchronized IMU. In addition, external ground truth poses of the drone are provided by a laser tracker system at 20Hz. However, partial tracking failures during high-acceleration maneuvers are reported. 
Lee et al. \cite{vividICRA2019} present the dataset ViViD, which contains sequences for visual navigation in poor illumination conditions. In addition to lidar and RGB-D data, they record thermal images with a resolution of 640x512 pixels at 20Hz. They provide ground truth poses from a MoCap system indoors and use a state-of-the-art lidar SLAM system for outdoor ground truth poses. However, the monocular DAVIS stream features only a resolution of 240x180 pixels and the timestamps between the sensors are not synchronized in hardware.

The dataset MVSEC by Zhu et al. \cite{zhu2018mvsec} as well as the recent dataset DSEC by Gehrig et al. \cite{Gehrig21ral} are most closely related to our work. Both MVSEC and DSEC contain a lidar sensor for depth estimation, two frame cameras, two event cameras as well as a GNSS receiver. DSEC addresses the limitation of a small camera baseline in MVSEC targeting automated driving scenarios. TUM-VIE comes with a number of complementary advantages. First, our event camera provides 10 times more pixels than the DAVIS m346B in MVSEC and 3 times more pixels than the Prophesee Gen3.1 in DSEC. This can help to study the impact of high resolution data on algorithmic performance and in principle allows a more detailed reconstruction.
Second, our visual stereo cameras encompass a largely increased field of view compared to the VI sensors in MVSEC. Third, in contrast to both MVSEC and DSEC, we provide a photometric calibration for our visual cameras, which is beneficial for employing direct VO methods on the intensity images \cite{engel2018pami}. Fourth, we provide calibrated IMU noise parameters which are required for accurate probabilistic modeling in state estimation algorithms \cite{schubert2018vidataset}. The MVSEC dataset features sequences on a hexacopter, on a car at low speed (12 m/s), and on a motorcycle at high speed (12-36 m/s) totalling about 60 minutes. However, only about 6 minutes of handheld indoor recordings are provided. Similarly, the DSEC dataset contains 53 minutes of outdoor driving on urban and rural streets. Contrary to that, the TUM-VIE dataset provides extensive handheld and head-mounted sequences during walking, running and sports in indoor and outdoor scenarios and under various illumination conditions, totalling 48 minutes excluding the calibration sequences.

\section{Dataset}

\subsection{Sensor Setup}
Table \ref{tab:sensors} gives an overview of our sensors and their characteristics. The full setup including the attached infrared markers can be seen in Figure \ref{fig::setup}. We use the Prophesee GEN4-CD Evaluation Kit which includes a 1 megapixel event sensor in a robust casing and a USB 3.0 interface. The Prophesee GEN4 CD sensor can handle a peak rate of 1 Giga-event per second. In most sequences we have a mean event rate of a few Mega-events per second.

\begin{figure}[t]
      \centering
      \includegraphics[scale=0.055]{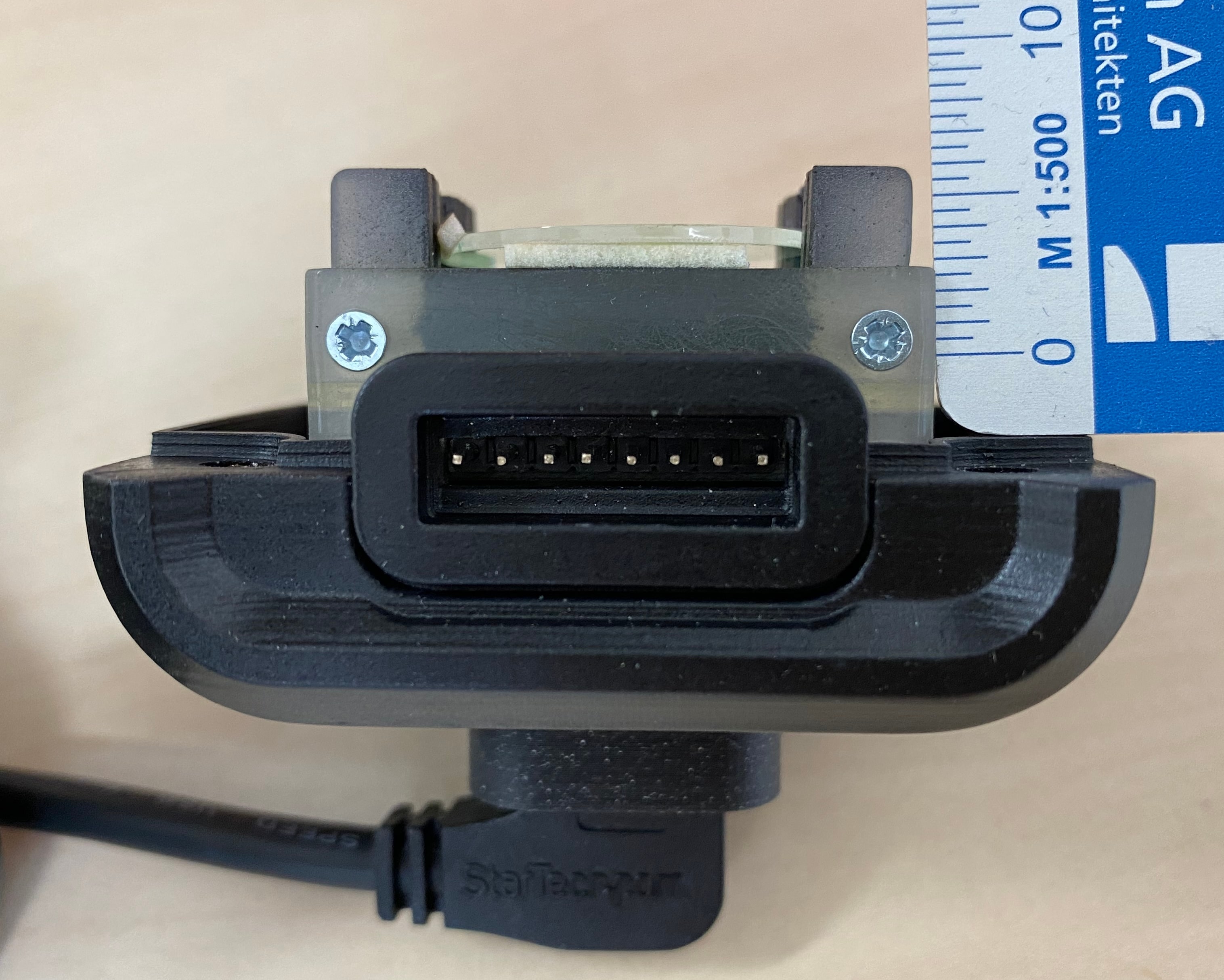}
      \caption{The infrared filter is mounted parallel to the sensor, at a distance of approximately 2 millimeter. The sensor plane is located approximately at the 5 millimeter mark of the ruler.}
      \label{fig::filter}
\end{figure}

The MoCap system emits 850nm infrared (IR) light in order to track the IR-reflective passive markers which are attached to the rigid body setup. Since the Prophesee GEN4 CD sensor is sensitive to IR light, moving the setup inside the MoCap room would create undesirable noise events. Therefore, we place an IR-blocking filter with cutoff frequency of 710nm parallel to the sensor, at a distance of approximately 2 millimeter, see Figure \ref{fig::filter}. The placement close to the sensor has the benefit that stray light inside the camera is also blocked.

\begin{table}[t]
\caption{Hardware Specification}
\label{tab:sensors}
\begin{center}
\tabcolsep=0.18cm
\begin{tabular}{c c c}

\hline
Sensor & Rate & Properties  \\
\hline
\begin{tabular}{c} 2x Prophesee GEN4-CD \\ {\scriptsize lens: Foctek M12-5IR}  \end{tabular}
&  $ \leq 10^9 \frac{\text{events}}{\text{s}}$
& \begin{tabular}{c}  1280x720 pixels \\  FOV: 90\textdegree H / 65\textdegree V \\ up to 124 dB  \\ \end{tabular} \\

\hline
\begin{tabular}{c} 2x IDS Camera uEye \\ UI-3241LE-M-GL \\ {\scriptsize lens: Lensagon BM4018S118} \end{tabular} 
& 20 Hz  
& \begin{tabular}{c} 1024x1024 pixels  \\ FOV:  101\textdegree H / 76\textdegree V \\ up to 60 dB \\ global shutter \end{tabular} \\

\hline
\begin{tabular}{c}IMU Bosch BMI160 \\ {\scriptsize integrated on Genuino 101} \end{tabular}  
& 200 Hz  
& \begin{tabular}{c} 3D accelerometer \\ 3D gyroscope \\ up to 60 dB \\ temperature \end{tabular} \\

\hline
MoCap OptiTrack Flex13 
& 120 Hz  
& 
\begin{tabular}{c} accurate 6D pose \\ 850nm IR light \end{tabular} \\

\hline
\end{tabular}
\end{center}
\end{table}

\subsection{Sequence Description}
Table \ref{tab:sequences} gives an overview of the available sequences. Total recording length amounts to 48 minutes excluding the calibration sequences. For each day of recording, we provide a sequence called \textbf{calib} which is used to calibrate the extrinsic and intrinsic parameters of all cameras. The sequence called \textbf{imu-calib} is used to obtain the transformation between the marker coordinate system (infrared markers for MoCap) and the IMU, as well as to determine the IMU biases and scaling factors. Furthermore, we provide two sequences called \textbf{calib-vignette}, which are used to compute the photometric calibration of the visual cameras.

\begin{table}[t]
\caption{Sequence Overview}
\label{tab:sequences}
\begin{center}
\renewcommand{\arraystretch}{1.2}
\begin{tabular}{c c c c}

\hline
Sequence  & Duration(s) & MER($10^6$ events/s)   \\ %

\hline
mocap-1d-trans  & 36.6  & 14.8 \\  
mocap-3d-trans & 33.2 & 24.9 \\   
mocap-6dof  & 19.5 & 27.15 \\     
mocap-desk & 37.5 & 29.7 \\ 
mocap-desk2 & 21.4 & 28.4 \\ 
mocap-shake  & 26.3 &  26.55  \\  
mocap-shake2  & 26.7 & 22.3 \\ 

\hline
office-maze & 160 & 28 \\   
running-easy & 73 & 27.25 \\ 
running-hard & 72 & 26.2 \\ 
skate-easy & 79 & 26.25 \\  
skate-hard & 86 & 25.7 \\ 

\hline
loop-floor0 & 284 & 29.55 \\ 
loop-floor1  & 257 & 29.55 \\ 
loop-floor2  & 240 & 28.3 \\ 
loop-floor3  & 256 & 29.7 \\ 
floor2-dark & 152 & 17.45 \\ 

\hline
slide & 196  & 28 \\

\hline
bike-easy  & 288 & 26.8 \\ 
bike-hard & 281 & 26.9 \\ 
bike-dark  & 261 & 25.35 \\ 

\hline
\end{tabular}
\end{center}
\end{table}

The first seven sequences in Table \ref{tab:sequences} with prefix \textbf{mocap} contain ground truth poses for the whole trajectory: \textbf{mocap-1d-trans} contains a simple one dimensional translational motion, \textbf{mocap-3d-trans} contains a three dimensional translation and \textbf{mocap-6dof} contains a full 6 DOF motion. All three sequences show a table containing diverse objects such as books, multiple similarly looking toy figures and cables. Most of the the scene is bounded by a calibration pattern placed behind the table. The sequences \textbf{mocap-desk} and \textbf{mocap-desk2} show a loop motion around two different office desks: \textbf{mocap-desk} shows two computer screens, a keyboard with some cables and the scene is bounded by a close-by white wall; \textbf{mocap-desk2} also shows two screens but the depth is less strictly bounded and there are also multiple calibration patterns and desk accessories visible. 

The sequence \textbf{office-maze} is recorded during a walk through various offices and hallways in the university building. The sequence \textbf{running-easy} is recorded in handheld mode while running through the corridor of the office. In the sequence \textbf{running-hard}, the camera is rapidly rotated during running such that it faces the office wall for short moments. This makes it hard to perform camera tracking. However, the wall features research posters such that there is still texture present for tracking.

The sequence \textbf{skate-easy} traverses the same corridor as running-easy and running-hard with a skateboard. In \textbf{skate-hard}, the camera is rapidly rotated to face the wall for a few seconds during the ride.

The sequence \textbf{loop-floor0} to \textbf{loop-floor3} are obtained by walking through the university building on the respective floor. These four datasets can be used to test loop closure detection algorithms in the event stream, which is still an immature research field. We additionally provide the sequence \textbf{floor2-dark}. This sequence shows the high dynamic range advantage of the event camera over the visual camera. We believe that with better event-based SLAM algorithms, the path could be tracked accurately even in this low-light condition, whereas state-of-the art visual SLAM systems fail, see Table \ref{tab:slamEval}.

\begin{figure}[t!]
     \centering
     \begin{subfigure}[b]{0.23\textwidth}
         \centering
         \includegraphics[width=\textwidth]{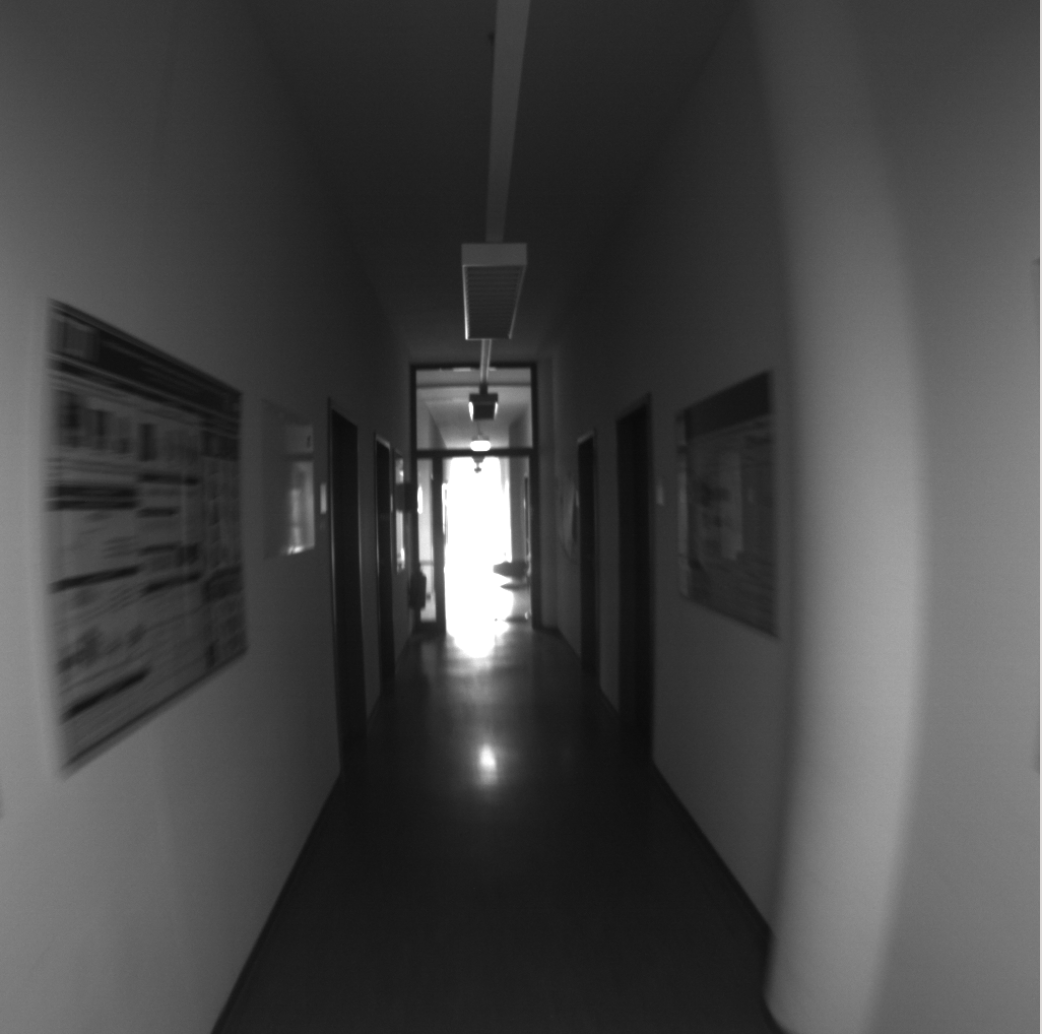}
         \caption{left visual frame}
         \label{fig:runeasyc0}
     \end{subfigure}
     \hfill
     \begin{subfigure}[b]{0.23\textwidth}
         \centering
         \includegraphics[width=\textwidth]{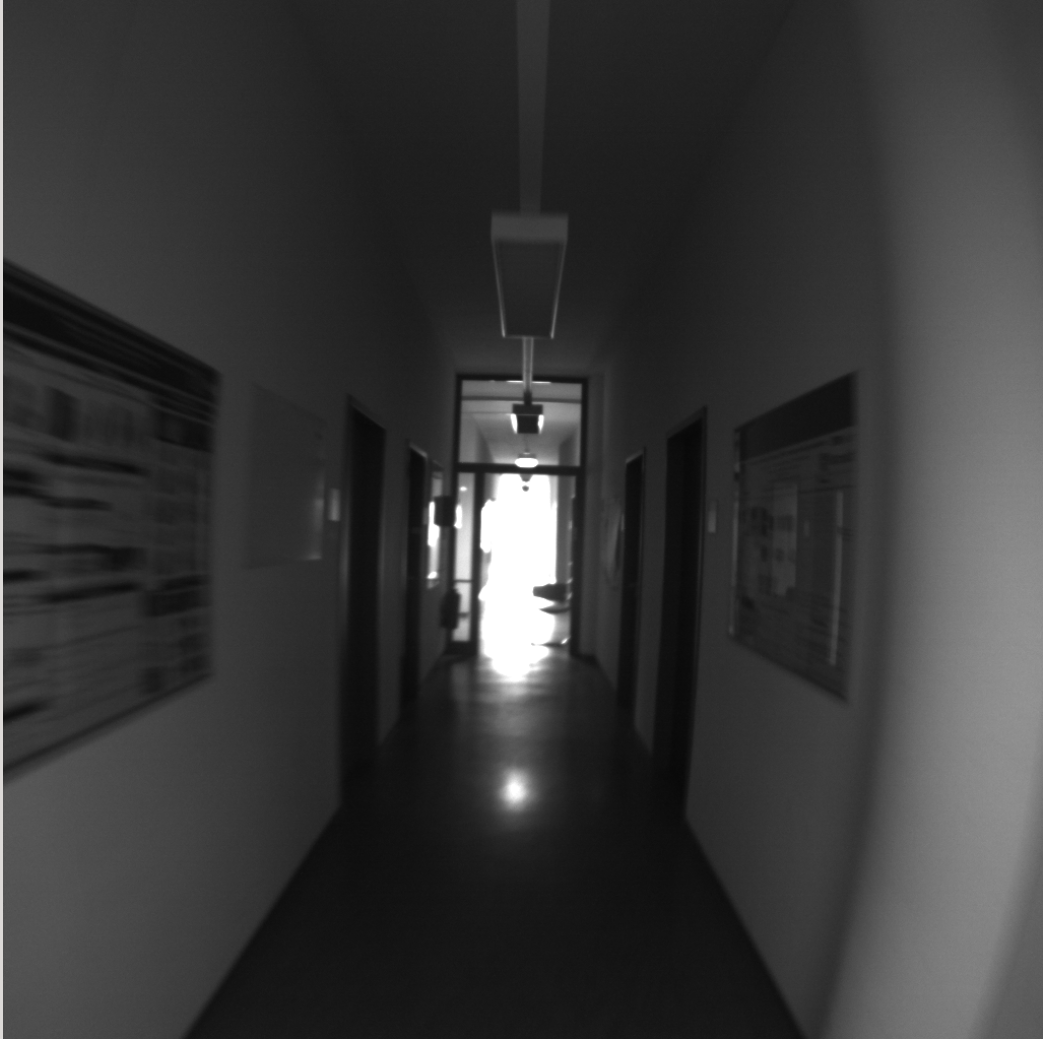}
         \caption{right visual frame}
         \label{fig:runeasyc1}
     \end{subfigure}
     \vskip\baselineskip
     \begin{subfigure}[b]{0.23\textwidth}
         \centering
         \includegraphics[width=\textwidth]{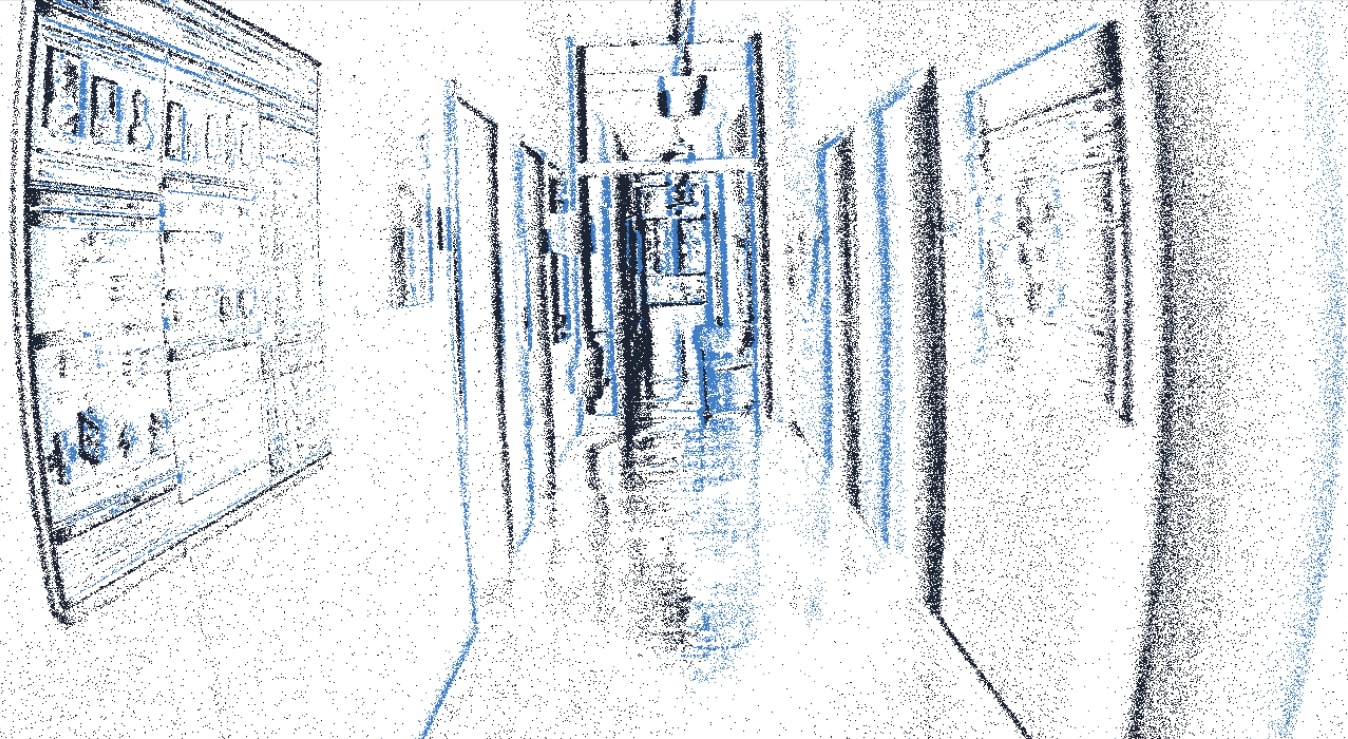}
         \caption{left event frame}
         \label{fig:runeasyc2}
     \end{subfigure}
     \hfill
     \begin{subfigure}[b]{0.23\textwidth}
         \centering
         \includegraphics[width=\textwidth]{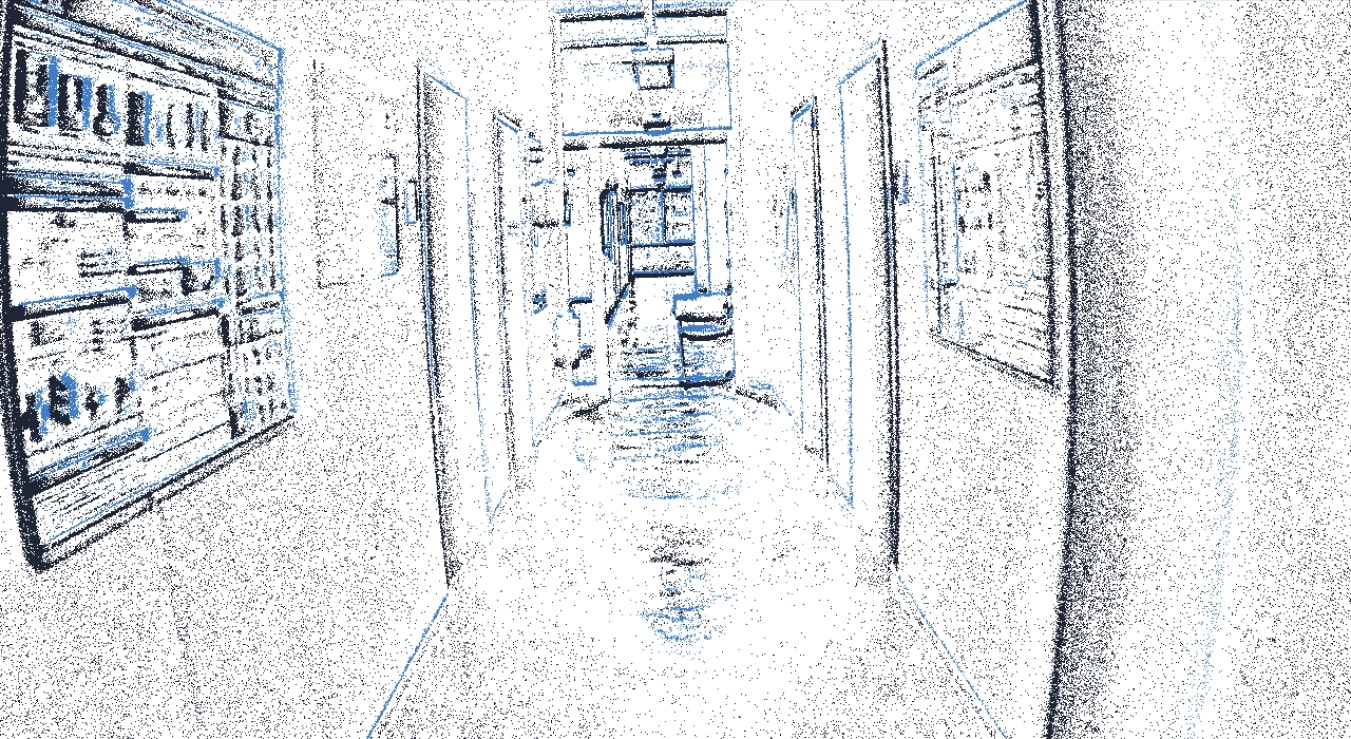}
         \caption{right event frame}
         \label{fig:runeasyc3}
     \end{subfigure}
        \caption{\textbf{run-easy} sequence at 25.1 seconds. The visual frames contain significant motion blur on the poster and oversaturation in the middle part of the image, whereas the event frames can capture many details.}
        \label{fig::runEasyScreenshots}
\end{figure}

\begin{figure}[t!]
     \centering
     \begin{subfigure}[b]{0.23\textwidth}
         \centering
         \includegraphics[width=\textwidth]{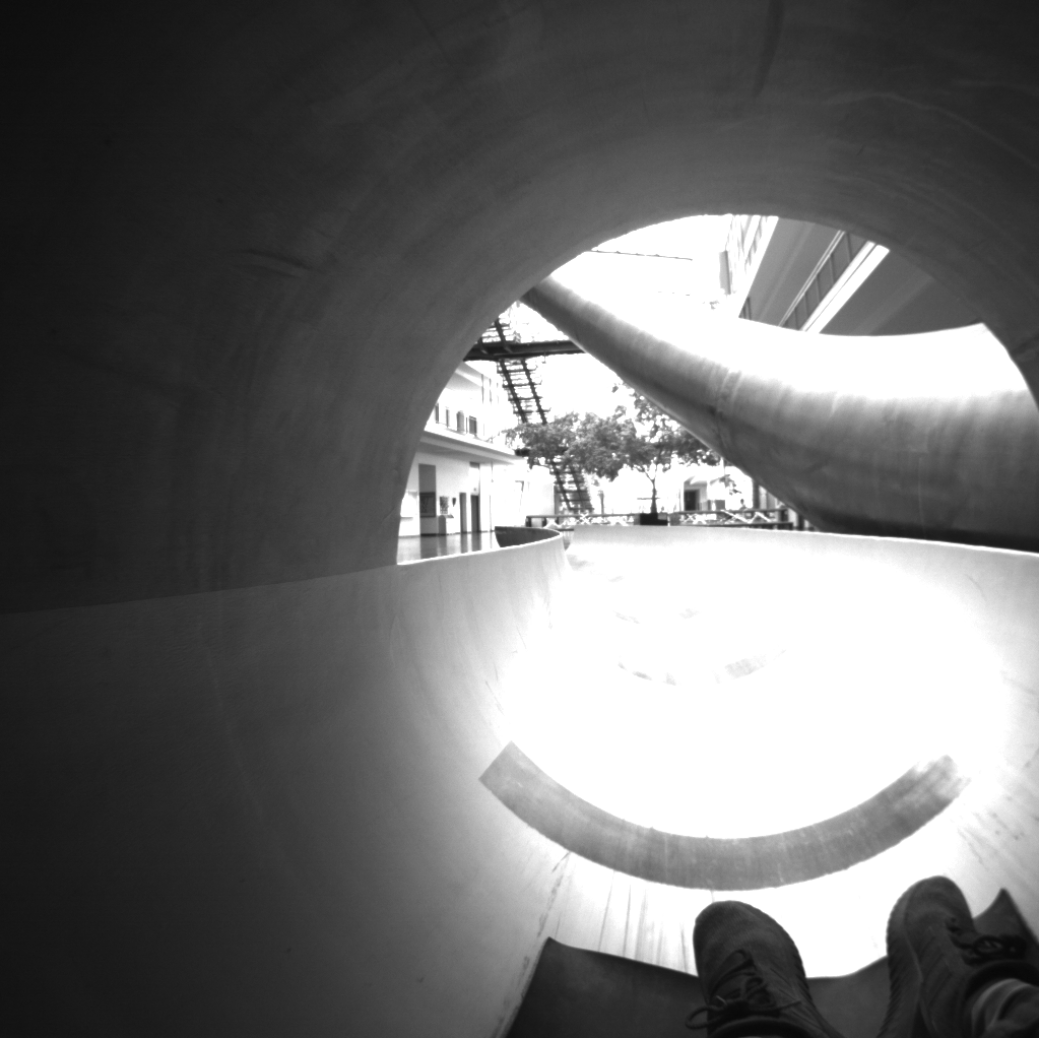}
         \caption{left visual frame}
         \label{fig:slidec0}
     \end{subfigure}
     \hfill
     \begin{subfigure}[b]{0.23\textwidth}
         \centering
         \includegraphics[width=\textwidth]{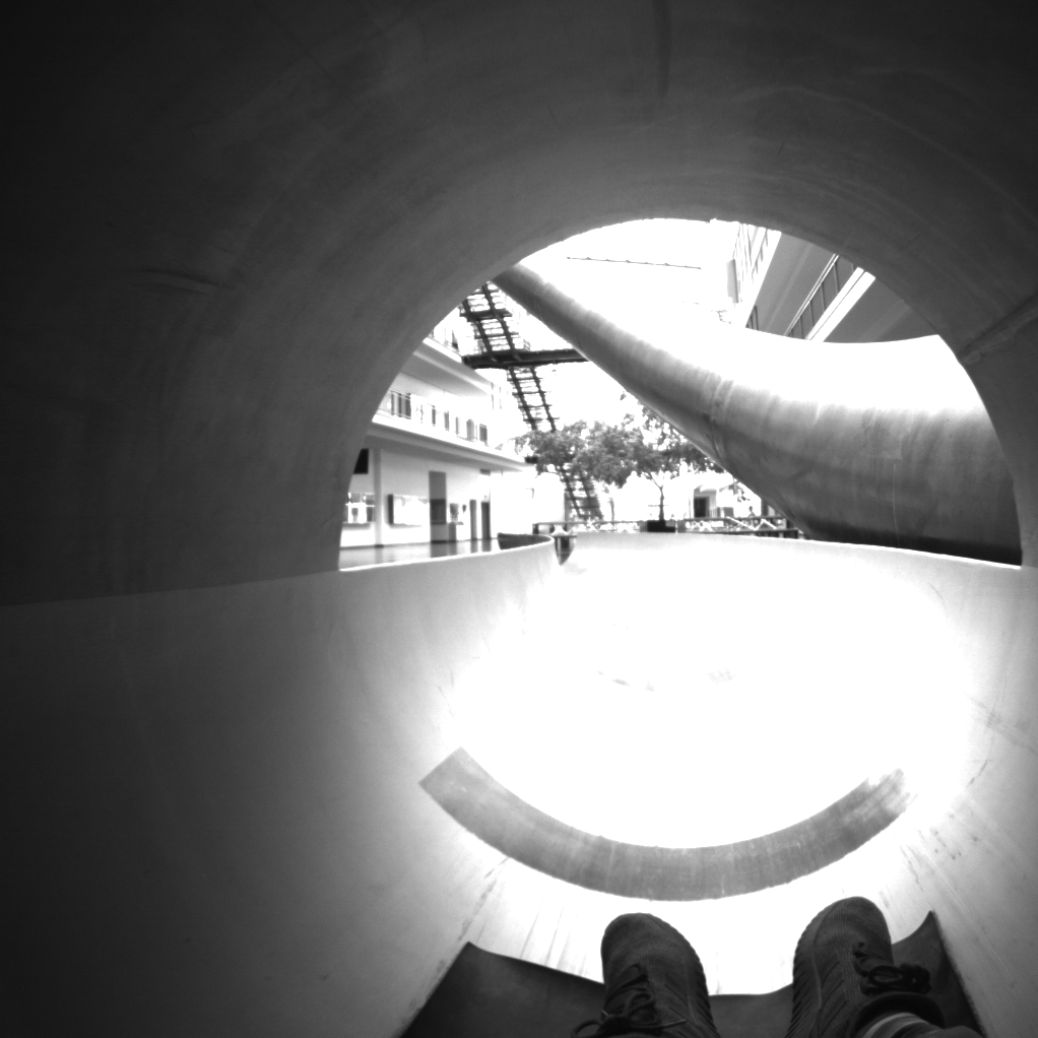}
         \caption{right visual frame}
         \label{fig:slidec1}
     \end{subfigure}
     \vskip\baselineskip
     \begin{subfigure}[b]{0.23\textwidth}
         \centering
         \includegraphics[width=\textwidth]{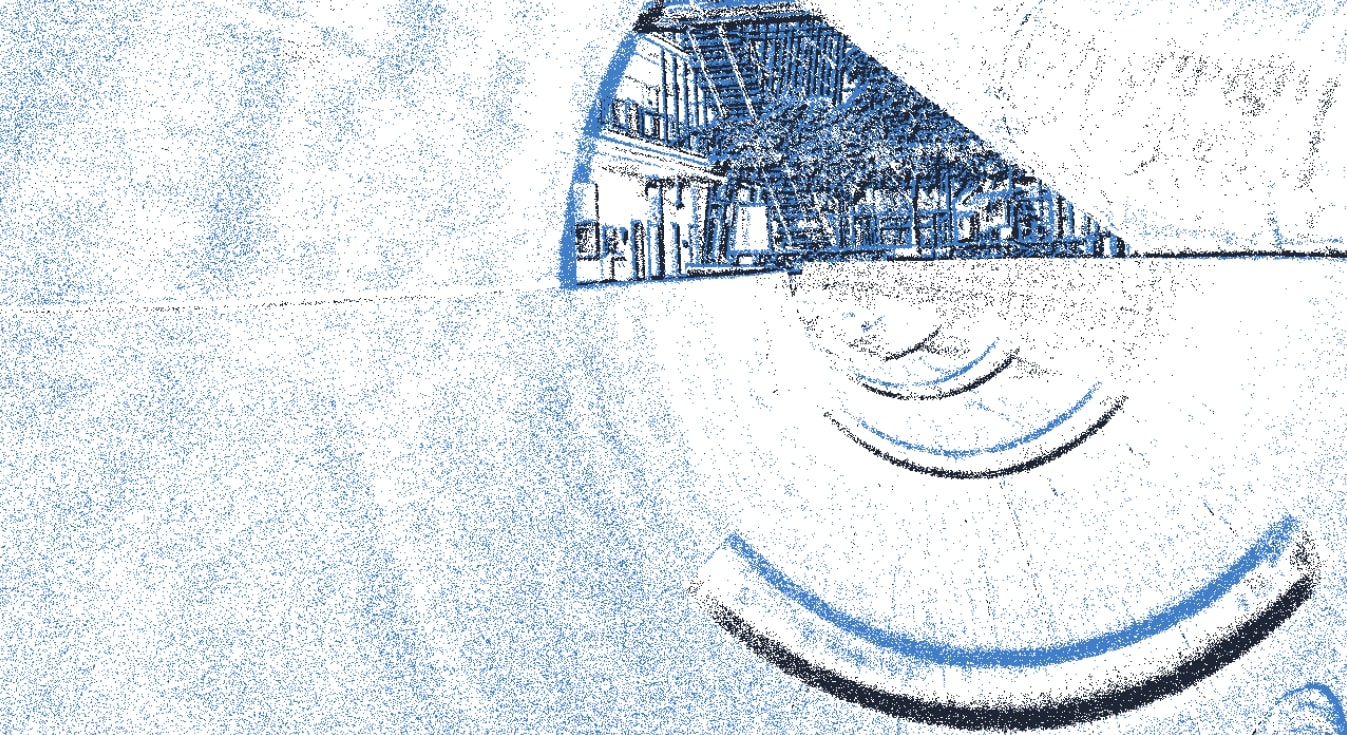}
         \caption{left event frame}
         \label{fig:slidec2}
     \end{subfigure}
     \hfill
     \begin{subfigure}[b]{0.23\textwidth}
         \centering
         \includegraphics[width=\textwidth]{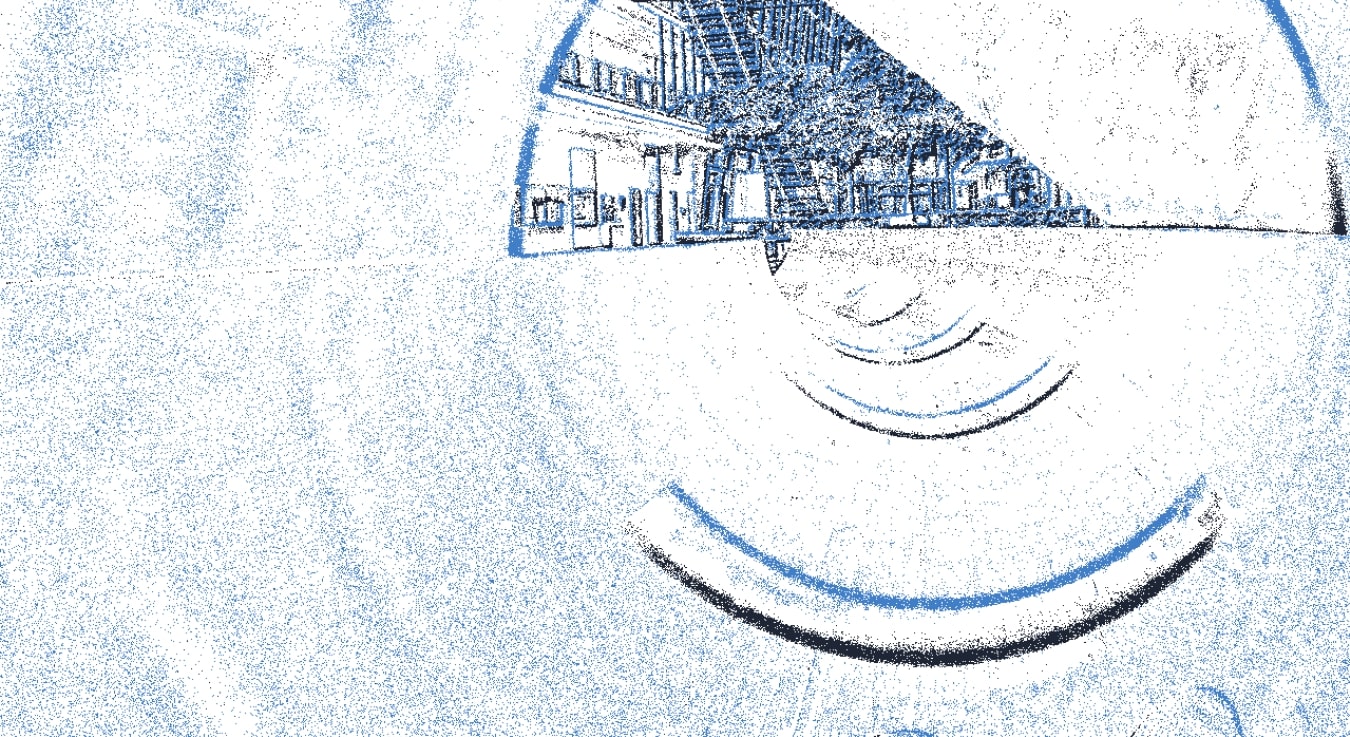}
         \caption{right event frame}
         \label{fig:slidec3}
     \end{subfigure}
        \caption{\textbf{slide} sequence at 93.7 seconds. Due ot the auto-exposure algorithm in the visual cameras, the exposure adapts quickly and details in the bright background are visible. However, a few frames earlier in this sequence, the frames are similarly oversatured as in Figure \ref{fig::runEasyScreenshots}.}
        \label{fig::slideScreenshots}
\end{figure}

The sequence \textbf{slide} shows a path through the university building while sliding from floor3 to floor0 in the middle of the sequence. This sequence contains high dynamic range and high speed motion.

The bike sequences are recorded with the helmet worn on the head, whereas all other sequences are recorded in handheld mode. \textbf{bike-easy} contains a biking sequence during the day at medium speed. \textbf{bike-hard} traverses the same path as bike-easy, additionally containing rapid rotations of the sensor setup, suddenly facing sideways or down on the road. \textbf{bike-dark} contains a slightly shorter path than the other bike sequences, recorded at night in low-light conditions outside.

In Figure \ref{fig::bikeNightScreenshots}, \ref{fig::runEasyScreenshots} and \ref{fig::slideScreenshots} we show both visual frames and events of a few selected sequences. The events are visualized as accumulated frames with an accumulation time of 5 milliseconds. Positive events are visualized in blue, negative events in black and white color is used to indicate no change.

\begin{figure}[t]
      \centering
      \includegraphics[scale=0.25]{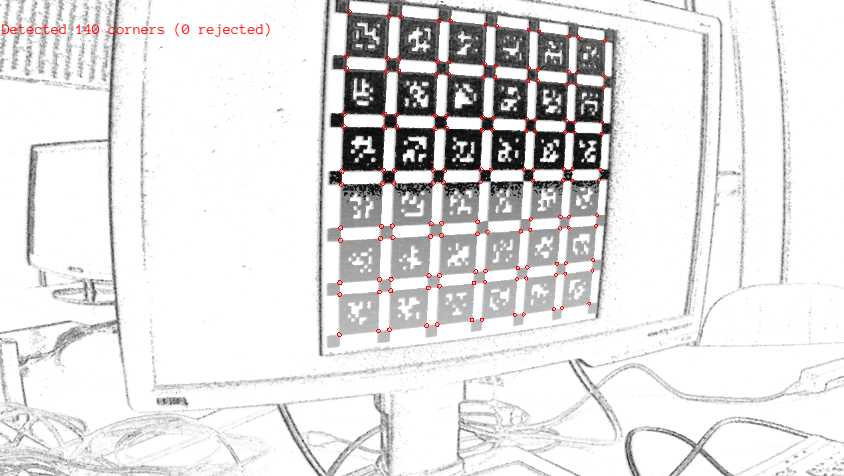}
      \caption{The April-grid pattern has been reconstructed based on creating the time surface. Using time surfaces showed slightly better AprilTag detection recall compared to accumulated event frames.}
      \label{fig::aprilgrid_ts}
\end{figure}

\section{Calibration}
\subsection{Intrinsic and Extrinsic Camera Calibration}

\begin{figure}[b]
      \centering
      \includegraphics[trim={0 0.5cm 0 0},clip,scale=0.85]{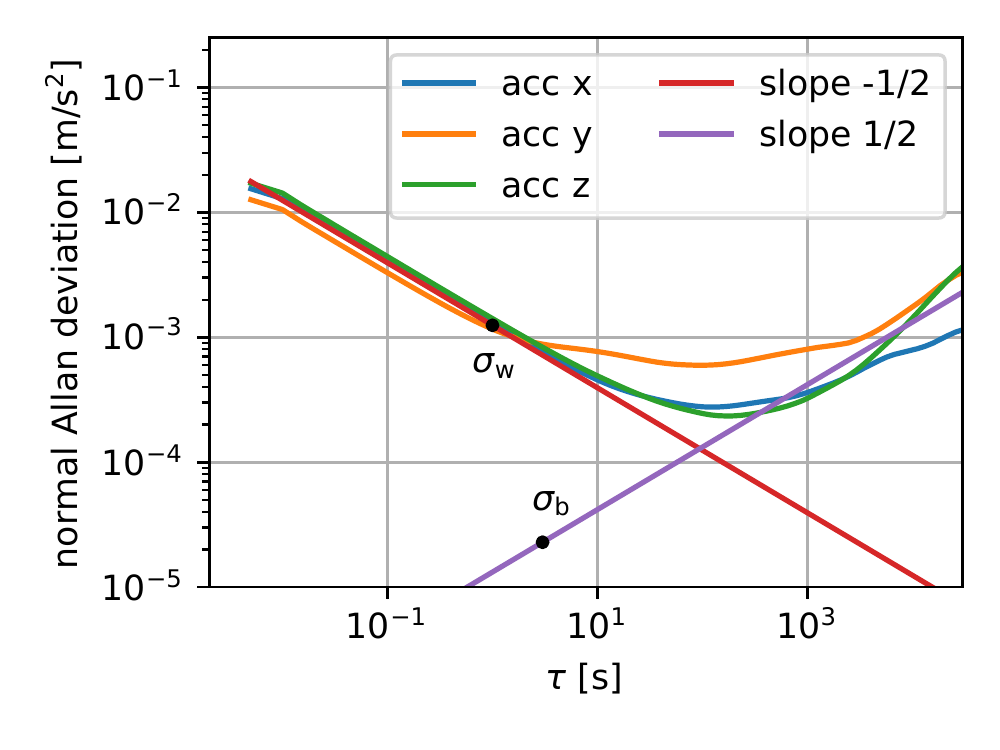}
      \caption{Accelerometer log-log plot of Allan deviation over integration time $\tau$ . We obtain the IMU noise values of $\sigma_{w, acc} = 0.0012$ at $\tau = 1$ and $\sigma_{b, acc} =  2.2856 \, \text{x} \, 10^{-5}$ at $\tau = 3$.}
      \label{fig::allan-a}
\end{figure}

The field of view among all cameras is only partially overlapping. Hence, to allow for extrinsic calibration between our camera modalities we use a pattern of AprilTags \cite{olson11AprilTags}. This pattern allows to perform robust data association between recognized tags from different cameras. It is displayed by a commodity computer screen, blinking at a fixed frequency. We move the multi-camera setup around the screen to capture the pattern from various poses. 

The intrinsic and extrinsic parameters for both visual and event cameras are determined in a joint optimization. To allow for accurate intrinsic parameter estimation, in particular radial distortion of the lenses, we validate for each calibration sequence that the detections of AprilTags are spread over the full image dimension of each camera.

\begin{figure}[b]
      \centering
      \includegraphics[trim={0 0.5cm 0 0},clip,scale=0.85]{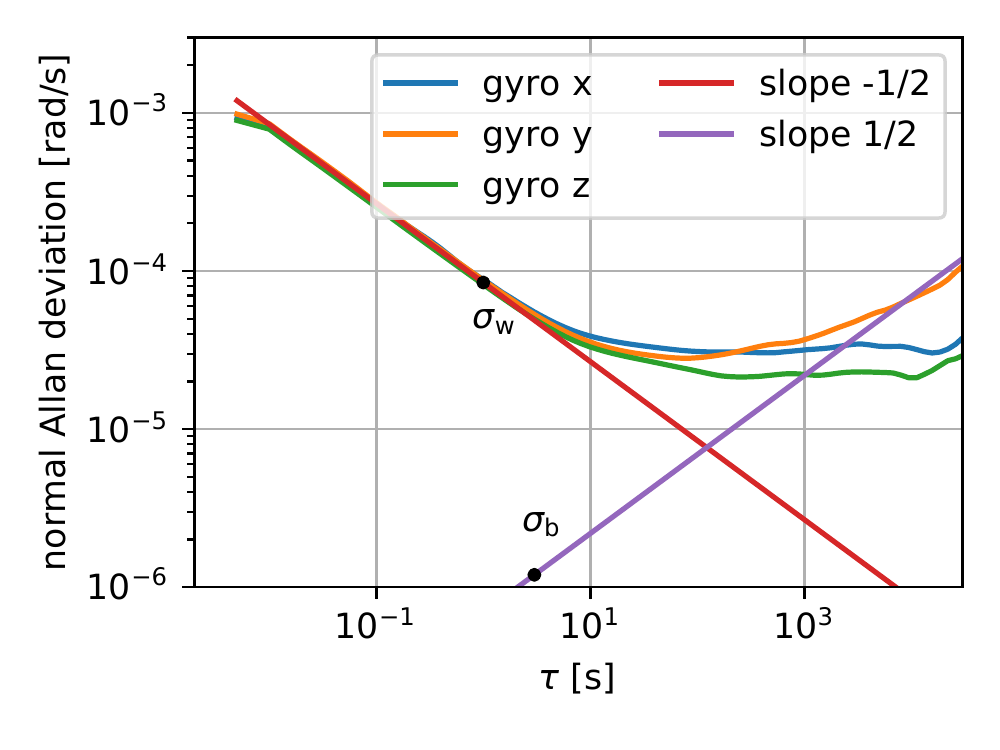}
      \caption{Gyroscope log-log plot of Allan deviation over integration time $\tau$. We obtain the IMU noise values of $\sigma_{w, gyro} = 8.4503 \, \text{x} \, 10^{-5}$ at $\tau = 1$ and $\sigma_{b, gyro} =  1.1961 \, \text{x} \, 10^{-6}$ at $\tau = 3$.}
      \label{fig::allan-g}
\end{figure}

To enable detection of AprilTags in the event stream, we transform the events into a frame-like structure. One popular method is to accumulate events within a certain time interval $t\in[t_{target} - \Delta t, t_{target} + \Delta t]$ into frames. However, in order to create sharper event frames at time $t_{target}$, we use the time surface representation instead. Inspired by \cite{zhou20esvo}, we use time surfaces which take the latest as well as the next future event at time $t_{target}$ into account. The time-surface $\mathcal{T}_{t_{target}}$ is created as follow, 
$$
\mathcal{T}_{t_{target}}[x,y] = \frac{1}{2} ( e^{-\alpha|t_{last} - t_{target}|} + e^{-\alpha|t_{next} - t_{target}|} ) \eqno{(2)}
$$
where $t_{last}$ is the timestamp of the previous event at pixel $[x,y]$ before $t_{target}$, $t_{next}$ is the timestamp of next event at pixel $[x,y]$ after $t_{target}$ and $\alpha$ is the decay rate. Using a time surface representation, we noticed a higher detection recall of AprilTags and hence a slightly more robust calibration. To perform this joint calibration, we modified the calibration tools from Basalt \cite{usenko19nfr}. A time surface created with the described method can be seen in Figure \ref{fig::aprilgrid_ts}. We provide the calibration parameters as well as the raw calibration sequences for each day of recording.

\subsection{IMU intriniscs, IMU-Camera-extrinsics}
Similar to the TUM-VI dataset \cite{schubert2018vidataset}, we assume that the IMU measurements are subject to white noise with standard deviation $\sigma_{w}$ and an additive bias value. The bias value is changing according to a random walk over time. The random walk is modelled as integration of white noise with standard deviation $\sigma_{b}$. To determine the parameter $\sigma_{w}$, a slope of $-\frac{1}{2}$ is fitted to the range of $0.02 \leq \tau \leq 1$ in the log-log plot of the Allan deviation. To determine the bias parameter $\sigma_{b}$, a slope of $\frac{1}{2}$ is fitted to the range $1000 \leq \tau \leq 6000$ seconds. The values for $\sigma_{w}$ and $\sigma_{b}$ can then be taken as the y-value of the fitted line at $\tau = 1$ and $\tau = 3$, respectively. This procedure is visualized in Figure \ref{fig::allan-a} and \ref{fig::allan-g}.

The extrinsic calibration between IMU and visual cameras is obtained by using a static Aprilgrid pattern. Calibration data is found in \textbf{imu-calib}. All six degrees of freedom are excited during this calibration while the pattern is constantly in the field of view of the visual cameras. The exposure time is set to a low value of 3.2 milliseconds to minimize motion blur but still allow detection of AprilTags. We leave the extrinsic between the visual cameras fixed and only optimize the transformation between IMU and visual camera frame.

\subsection{Temporal Calibration}

\begin{figure}[t]
      \centering
      \includegraphics[scale=0.35]{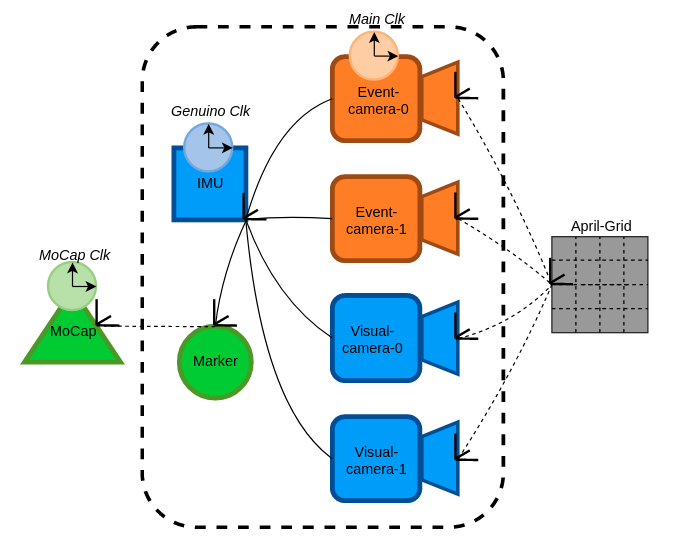}
      \caption{Coordinate systems which we have used in this paper. Solid lines are the relative poses found during calibration, dashed lines are the relative poses measured in real-time. Colors indicate the respective raw sensor clock before temporal calibration.}
      \label{fig::ref-frames}
\end{figure}

All sensors on the setup are synchronized in hardware. An overview of the different sensor clocks can be seen in Figure \ref{fig::ref-frames}. We report all timestamps in the left event camera's clock (main clock). The synchronization between left and right event camera is achieved through the available synchronization connections of the Evaluation Kit.

The Genuino 101 triggers the visual cameras, measures exposure start and stop timestamps and reads the IMU values. As mentioned above, we define the timestamp of an image as mid-exposure point. Hence, timestamps of the images  are well-aligned with IMU measurements in Genuino clock. However, due to the IMU readout delay, there exits a constant offset of 4.77 milliseconds. We determine this offset exactly once during calibration and correct it in all recordings.

To transform IMU and image timestamps from Genuino clock to main clock, we additionally measure exposure start and stop timestamps via trigger signal in the left event camera. Data association between triggers on the Genuino and triggers on the event camera is achieved through a distinctive startup sequence, which alternates extremely high and low exposure times. We notice a linear relation between Genuino and main clock, i.e. there exists a constant offset $b$ and a small constant clock drift $m$ between both clocks. Whereas the value of $b$ is different for each sequence, the value of $m$ is usually around 2 milliseconds per minute (Genuino clock runs faster than main clock). We obtain the linear coefficients from a least-squares fit and correct for them in each sequence, such that all sensor data is provided in main clock.

\begin{figure}[t]
      \centering
      \includegraphics[scale=0.45]{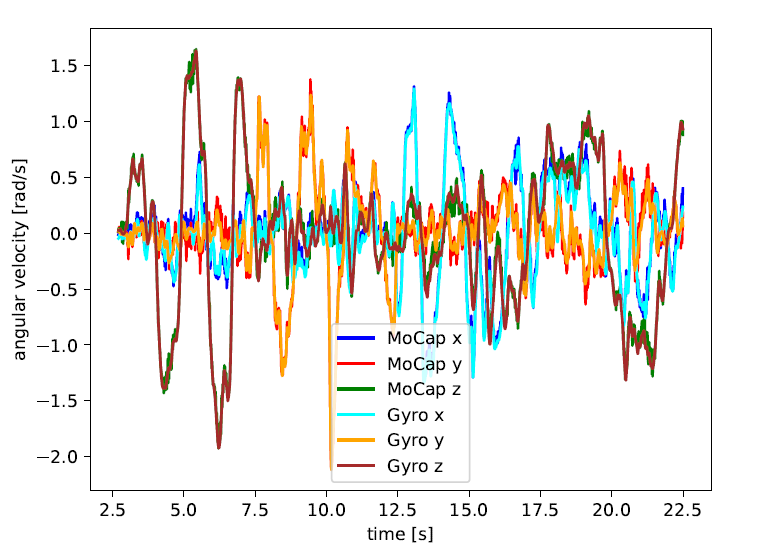}
      \caption{Alignment of angular velocities between gyroscope measurements and Mocap estimations (using central differences of poses). The obtained time offset $\Delta t_{gt,m}$ between MoCap clock and IMU clock is provided for each sequence.}
      \label{fig::mocapAlign}
\end{figure}

Ground truth poses are recorded in MoCap clock. We model a constant offset $\Delta t_{gt,m}$ between MoCap clock and main clock. Since all data is recorded on the same computer, a rough estimate of this offset is obtained from the system's recording time. A refinement of this estimate is obtained by aligning angular velocities measured by the IMU with estimated angular velocities which are obtained from central differences between MoCap poses, as can be seen in Figure \ref{fig::mocapAlign}. The offset $\Delta t_{gt,m}$ which minimizes the sum of all squared errors is estimated for each sequence. For the trajectories where we exit and later re-enter the MoCap room, we compute this offset once for the beginning and once for the end of the sequence to account for clock drift. The provided ground truth poses in the dataset have corrected timestamps and are reported in the main clock. Additionally, we provide the estimated offsets to facilitate custom time-alignment approaches.

\subsection{Biases and Event Statistics}
Table \ref{tab::biases} shows the biases of the event cameras, which are the same in all sequences. The ratio of ON over OFF pixels is approximately 0.7 in all sequences and for both cameras. 

\begin{table}[b]
\caption{Biases of Prophesee GEN4-CD sensor}
\label{tab::biases}
\begin{center}
\begin{tabular}{c c}
bias name & value   \\
\hline
bias\_diff & 69  \\
bias\_diff\_off & 52  \\
bias\_diff\_on & 112 \\
bias\_fo\_n & 23   \\
bias\_hpf & 48  \\
bias\_pr & 151  \\
bias\_refr & 45 \\
\end{tabular}
\end{center}
\end{table}

\section{Evaluation of stereo-VIO Algorithms}

\begin{table}
\caption{Absolute trajectory errors (ATE) [m]}
\label{tab:slamEval}
\begin{center}
\begin{tabular}{c c c c}
\hline
Sequence  & Basalt    &  VINS-Fusion  & length[m]\\
\hline
mocap-1d-trans  &  0.003 &  0.011  &  5.01\\ 
mocap-3d-trans  &  0.009 &  0.010  &  6.85\\ 
mocap-6dof      &  0.014 &  0.017  &  5.30\\
mocap-desk      &  0.016 &  0.058  &  9.44\\ 
mocap-desk2     &  0.011 &  0.013  &  5.34\\ 
mocap-shake    &  x     &  x  &  24.5\\    
mocap-shake2    &  x     &  x  &  26.2\\     
\hline
office-maze   &  0.64  &  4.40   &  205\\  
running-easy  &  1.34   &  0.78  &  113\\ 
running-hard  &  1.03   &  1.74   &  117\\  
skate-easy    &  0.22  &  1.74   &  114\\ 
skate-hard    &  1.78   &  0.97  &  113\\ 
\hline
loop-floor0  &  0.58 &  3.43 &  358\\  
loop-floor1  &  0.66 &  1.72 &  338\\  
loop-floor2  &  0.48 &  2.04 &  282\\  
loop-floor3  &  0.51 &  8.36 &  328\\  
floor2-dark  &  4.54  &  4.54 &  254\\  
\hline
slide  &  1.54  &  2.44  &  248\\ 
\hline
bike-easy  &  1.62 &  13.10  &  788\\ 
bike-hard  &  2.01 &  9.88  &  784\\  
bike-dark  &  6.26 &  20.20  &  670\\ 
%
\hline
\end{tabular}
\label{tab::ate}
\end{center}
\end{table}

\begin{figure}
     \centering
     \begin{subfigure}[b]{0.23\textwidth}
         \centering
         \includegraphics[width=\textwidth]{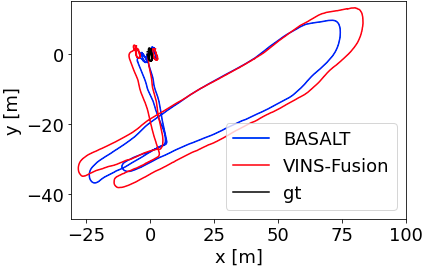}
         \caption{floor\_0}
         \label{fig:floor_0}
     \end{subfigure}
     \hfill
     \begin{subfigure}[b]{0.23\textwidth}
         \centering
         \includegraphics[width=\textwidth]{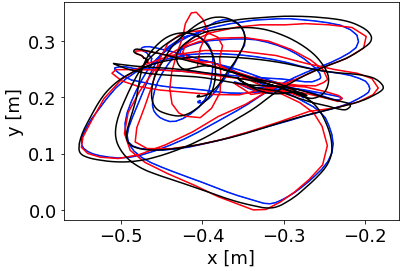}
         \caption{mocap\_6dof}
         \label{fig:mocap_6dof}
     \end{subfigure}
     \vskip\baselineskip
     \begin{subfigure}[b]{0.23\textwidth}
         \centering
         \includegraphics[width=\textwidth]{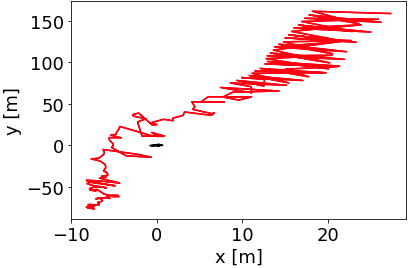}
         \caption{shake\_2}
         \label{fig:shake_2}
     \end{subfigure}
     \hfill
     \begin{subfigure}[b]{0.23\textwidth}
         \centering
         \includegraphics[width=\textwidth]{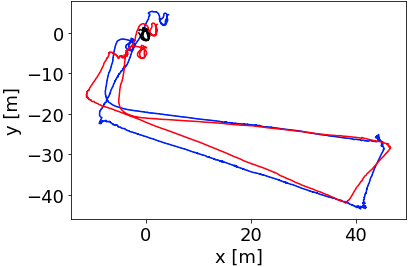}
         \caption{floor\_2\_night}
         \label{fig:floor_2_night}
     \end{subfigure}
     \vskip\baselineskip
     \begin{subfigure}[b]{0.23\textwidth}
         \centering
         \includegraphics[width=\textwidth]{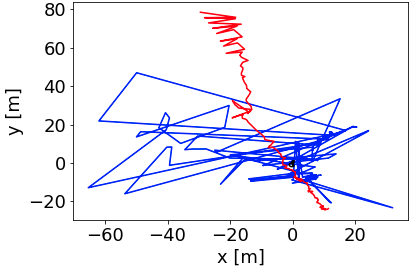}
         \caption{shake\_1}
         \label{fig:shake_1}
     \end{subfigure}
     \hfill
     \begin{subfigure}[b]{0.23\textwidth}
         \centering
         \includegraphics[width=\textwidth]{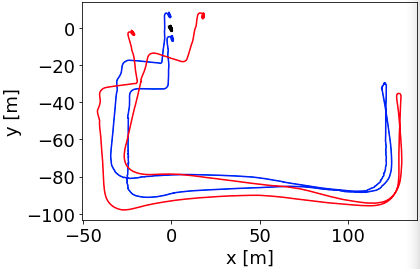}
         \caption{bike\_night}
         \label{fig:bike_night}
     \end{subfigure}
        \caption{Results of evaluated methods for 2 easy sequences (a, b) and 4 challenging sequences (c, d, e, f) from our dataset. The ground truth is shown in black for the segments of the trajectory where it is available. The presented results are obtained with default parameters. Noise parameters are set to inflated values from the Allan plots in Figure \ref{fig::allan-a} and  \ref{fig::allan-g} to account for unmodeled noise and vibrations.}
        \label{fig:paths_graphs}
\end{figure}

To verify the data and calibration quality, we test all sequences with state-of-the-art open-source visual-inertial odometry systems. We provide evaluation for Basalt \cite{usenko19nfr} and VINS-Fusion \cite{qin2017vins,qin2019a,qin2019b}. The methods are used with default parameters on full resolution images (1024 x 1024 pixels). The results are summarised in Table \ref{tab::ate} and a visualization for some sequences is showed in Figure \ref{fig:paths_graphs}. An x in Table \ref{tab::ate} indicates an ATE larger than the sequence's path length or that the method fails. Basalt and VINS-Fusion perform well for non-challenging sequences as expected.

However, our evaluation also shows that Basalt and VINS-Fusion result in large drift for most of the challenging sequences, e.g. with fast motion in \textbf{running-easy}, \textbf{running-hard}, \textbf{skate-hard}, \textbf{slide}, and \textbf{bike-hard} or with low light, e.g.\ in \textbf{floor2-dark} and \textbf{bike-dark}. This means that the dataset is challenging enough for state-of-the-art visual-inertial systems and can be used for further research in event-based visual-inertial odometry algorithms.

\section{Conclusion}
In this paper, we propose a novel dataset with a diverse set of sequences, including small and large-scale scenes. We specifically provide challenging sequences in low light and high dynamic-range conditions as well as during fast motion. Our dataset composes of a high-resolution event stream and images captured by a wide field of view lens, as well as IMU data and partial ground truth poses.

We evaluate our dataset with state-of-the-art visual-camera-based stereo-VIO. The results show that there are open challenges which need new algorithms and new sensors to tackle them. Event-based odometry algorithms are still immature compared to frame-based methods, which makes it difficult to present an evaluation of event-based algorithms.
Hence, our dataset can be useful for further
research in the development of event-based visual-inertial odometry, as well as 3D reconstruction and sensor fusion algorithms.


\section*{ACKNOWLEDGMENT}
We thank everybody who contributed to the paper. In particular, we want to thank Mathias Gehrig (University of Zurich) for supporting us with the h5 file format. We also thank Lukas Koestler (Technical University of Munich) for helpful suggestions and proofreading the paper. Additionally, we thank Yi Zhou (HKUST Robotics Institute) for his helpful comments. This work was supported by the ERC Advanced Grant SIMULACRON.


\balance
\printbibliography

\end{document}